\documentclass[conference]{IEEEtran}

\usepackage[utf8]{inputenc}
 \ifCLASSINFOpdf 
	\usepackage[pdftex]{graphicx}
	\DeclareGraphicsExtensions{.pdf,.jpeg,.png}
\else
	\usepackage[dvips]{graphicx}
	\DeclareGraphicsExtensions{.eps}
\fi
\usepackage[noadjust]{cite}
\usepackage{multicol}
\usepackage{amsmath}
\usepackage{algorithmic}
\usepackage{array}
\ifCLASSOPTIONcompsoc
	\usepackage[caption=false,font=footnotesize,labelfont=sf,textfont=sf]{subfig}
\else
	\usepackage[caption=false,font=footnotesize]{subfig}
\fi
\usepackage{fixltx2e}
\usepackage{url}
\begin{document}

\title{Client Profiling for an Anti-Money Laundering System}

\author{\IEEEauthorblockN{Alexandre, Claudio}
\IEEEauthorblockA{Faculdade de Ciências da Universidade de Lisboa\\BioISI-MAS\\Lisbon, Portugal\\
Email: calexandre@di.fc.ul.pt}
\and
\IEEEauthorblockN{Balsa, João}
\IEEEauthorblockA{Faculdade de Ciências da Universidade de Lisboa\\BioISI-MAS\\Lisbon, Portugal\\
Email: jbalsa@ciencias.ulisboa.pt}
}
\maketitle

\begin{abstract}
We present a data mining approach for profiling bank clients in order to support the process of detection of anti-money laundering operations. We first present the overall system architecture, and then focus on the relevant component for this paper. We detail the experiments performed on real world data from a financial institution, which allowed us to group clients in clusters and then generate a set of classification rules. We discuss the relevance of the founded client profiles and of the generated classification rules. According to the defined overall agent-based architecture, these rules will be incorporated in the knowledge base of the intelligent agents responsible for the signaling of suspicious transactions.
\end{abstract}

\begin{IEEEkeywords}
anti-money laundering; data mining; classification; customer clustering; multiagent systems; suspicious transactions
\end{IEEEkeywords}

\section{Introduction} 
Acts of prevention and fight against money laundering (ML) crimes are prioritized by almost every government in the world, at the same level of the most relevant global issues. Money laundering is a crime that typically consists in making a certain illegal financial gain into a legal gain. According to the United Nations Office on Drugs and Crimes (UNODC) the annual global estimate of laundered money is about 2\% - 5\% of the Gross World Product, or US\$800 billion - US\$2 trillion \cite{unodc2014}. As if the financial volume were not enough, another reason for governments to focus on this crime is for the fact that it is clearly connected to other types of crimes such as illegal drug trade, fraud, corruption, kidnapping, terrorism, arms smuggling, among others.\\

Most countries’ financial authorities, usually Central Banks, are responsible for controlling and defining anti-money laundering (AML) regulations, demanding from financial institutions the implementation of procedures that apply the defined norms. However, the constant increase on the amount of financial transactions along with the high frequency of publication of new national and international rules, cause the lack of efficiency and timing of the fight and prevention activity.\\

Institutions already use semi-automated processes to indicate suspicious ML transactions, based on medians and predetermined standard irregularities. Still, due to the sophistication of this criminal activity, the most critical part of process is still performed by human analysts. To make the identification and analysis of the suspect transactions more agile, using data mining techniques and intelligent agents, reducing the need of human intervention is the main goal of this doctoral project, as detailed in \cite{Alexandre15}.\\

In this paper, we focus on the results related to one of the systems components, that has to do with the identification of a financial institution’s customers’ behavior patterns, using the entire checking accounts transaction database of transactions made over a year. In section II, we provide some context for our work, presenting the system where the present research is used. In section III, we present some related work on the use of data mining techniques in financial applications.\\

\section{AML Multiagent System Proposed} 
\subsection{Problem Definition}
Money Laundering is characterized by a set of commercial or financial operations that aim to incorporate in each country’s economy, in a transitory or permanent way, illicitly obtained resources, goods or values.\\

The role of financial institutions is to find ways to identify, among the huge number of operations that occur every day, those suspicious transactions and then investigate them in more detail.\\

The motivation for using an intelligent agent based solution results from the analysis of the problem, as described in \cite{Alexandre15}, and, mainly, in the observation that some of the tasks that we want to automate (at least partially), match perfectly the principles behind multi-agent system definition \cite{Wooldridge09}. We need a set of entities (agents) with autonomy to accomplish specific tasks and that keep contact with other agents in order to reach a common objective. Every agent must have its own knowledge and be able to ponder and come to an intelligent decision. Besides, they need to present scalability and be flexible \cite{Demazeau95}.

\subsection{System Architecture}
\begin{figure}[!t]
\centering
\includegraphics[width=3.5in]{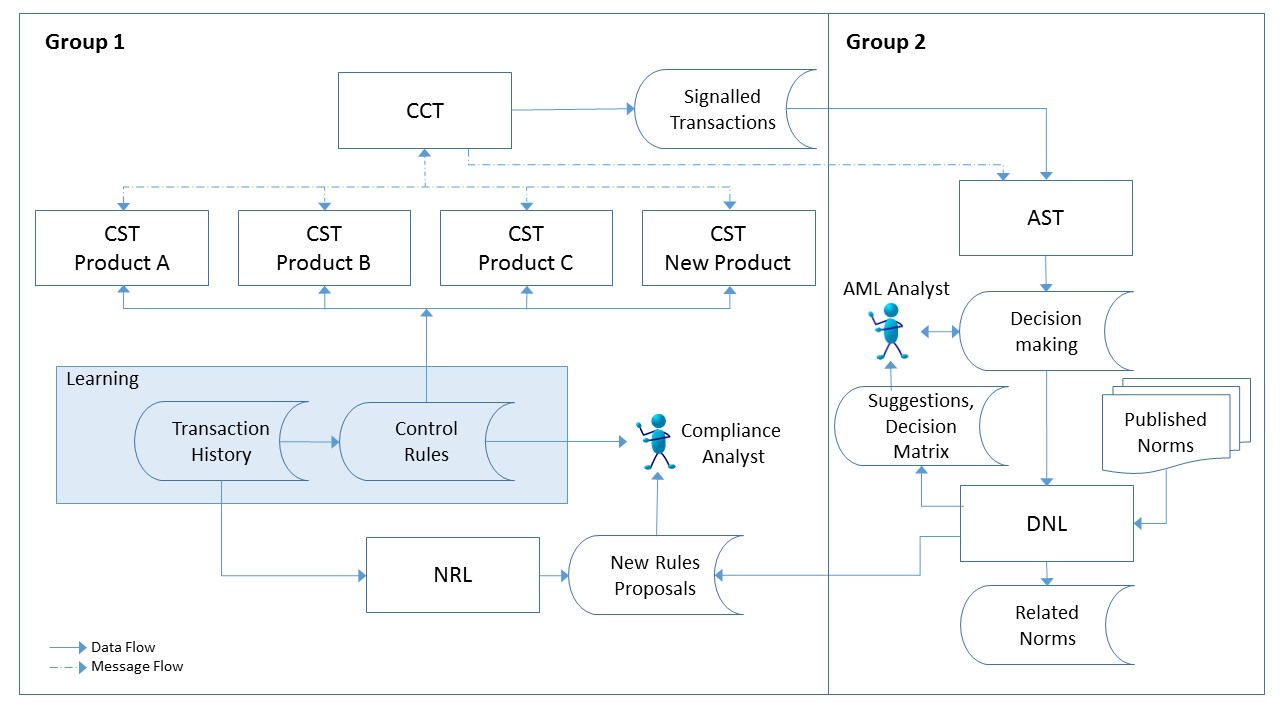}
\caption{General Architecture of the Proposed System}
\label{architecture}
\end{figure}

We consider two groups of agents, according to their role in the process. The first one is a group responsible for capturing suspicious transactions (CST), while the second is responsible for analyzing suspicious transactions (AST) identified by agents from the first group. Fig. \ref{architecture} shows a schematic view and the global architecture flow of the proposed system.\\

For the CST to work, it is necessary to know customers’ behavior patterns so that we can establish controlling rules. Other rules resulting from the defined norms made by the financial system regulator entities will be included in these controlling rules. Fig. \ref{architecture} also shows, in highlight, the step called learning that incorporates the method we describe in this paper.

\section{Related Work} 
From the work that has been done in the application of data mining technique to the finance domain, we select here some of the most relevant for our research.\\

Considering the data transformation process, which was especially critical in our work, in \cite{Zang03}, a discretization process was also applied to data set in order to find a more adequate set of clusters. Resources are mapped to n+2 dimensional Euclidean space, n being the customer’s attributes, one dimension for time and another for transactions. Customers’ transactions are projected according to time, accumulating transactions and its frequency to create a histogram. Clusters are created based on segments of the histogram. 
Analyses of local and global correlations are then applied to detect suspicious patterns. A good way to analyze individual behavior and/or group behavior is to examine their operations to detect suspicious behaviors related to abnormal peaks on the histogram. However, when it is necessary to analyze a big number of clients and transactions over a long period of time, it may become difficult to detect suspicious cases, since there might be few peaks or none at all in the histogram.\\

The implementation proposal in \cite{Kingdon04} is an SVM extension and a matrix with massive dimensionality was created. In theory, a positive factor of this approach is that it can handle heterogeneous data set, however, the matrix dimensionality left many doubts about the performance \cite{NhienAn10}.\\ 

In \cite{Tang05}, the authors propose an extension of the support vector machine (SVM) \cite{Scholkopf00} to detect customers’ abnormal behavior. A combination of an RBF kernel (radial basis function) is presented with improvements \cite{Scholkopf01} as a definition for distinct distances \cite{Wilson97} and supervised and unsupervised SVM algorithms. An SVM class \cite{Scholkopf00} is a supervised way of learning used to detect value anomalies in a group of data without classes. The advantage of this approach is that you can deal with heterogeneous datasets.\\

A combination of clustering and MLP (multilayer perceptron) was proposed by \cite{NhienAn09}. A simple center-based clustering technique is used to detect suspicious cases of money laundering. This technique is based on two main characteristics, which are then used as an MLP creation process entry. The preliminary results show that this approach is efficient. However, the number of characteristics and training patterns is too small and that could affect precision.\\

In \cite{NhienAn10}, the authors present a case study corresponding to the application of a knowledge base solution that combines data mining techniques, clustering, neural networks and genetic algorithms to detect money-laundering patterns.

\section{Data Mining Process} 
\begin{figure}[!t]
\centering
\includegraphics[width=3.4in]{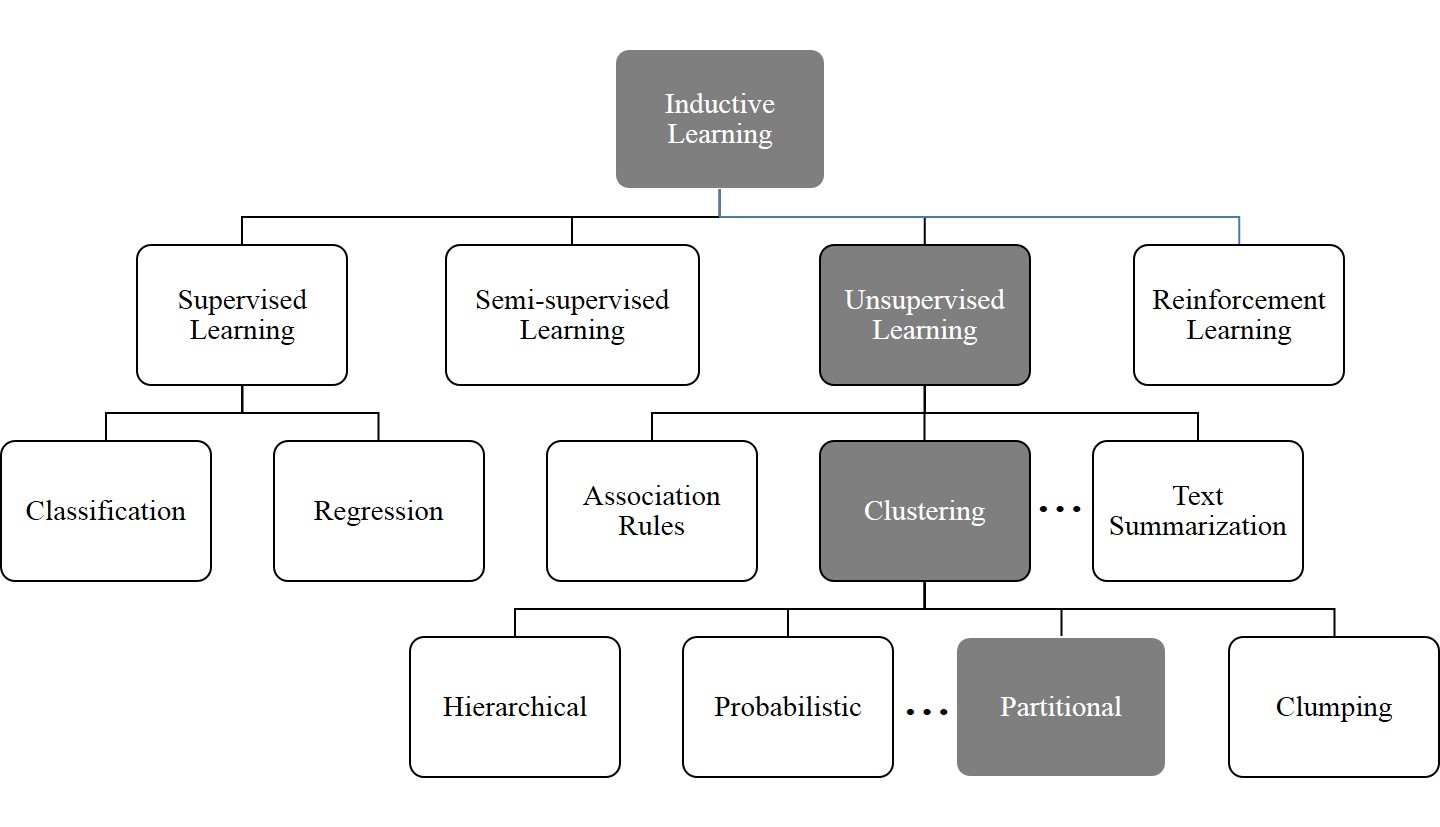}
\caption{Inductive Learning Hierarchy}
\label{hierarchy}
\end{figure}

Basically, what differentiates the types of learning is the existence or not of a class attribute docketing the registries of the data group used. When this class exists, the learning process will be supervised; when only part of the examples have this attribute it will be semi-supervised; and when these labels don’t exist at all the learning process is unsupervised \cite{Metz06}. Fig. \ref{hierarchy}  illustrates this hierarchy besides highlighting the way chosen in this paper.

\subsection{Choice of tools and environment definition}
The data used in this paper don’t have a class attribute (unsupervised). The initial goal is to discover patterns from any regular characteristic in the dataset (Clustering), trying to form groups of clients with similar characteristics and mutually exclusive (partitional).\\

Despite the classic problem associated with the K-means algorithm \cite{Hamerly02}, the necessity of defining in advance the number of clusters to be used, it is one of the most used methods, perhaps for its simplicity, efficiency and for being present in almost every platform that implements and automates the data mining process.\\

Another point that needs to be looked at when talking about K-means is that we obtain better results when dealing with continuous numerical attributes in comparison with nominal attribute use. The reason being that, originally, it uses squared Euclidean distance to calculate proximity. In nominal attributes this calculation can’t be made.\\

The easy of use of the WEKA tool (Waikato Environment for Knowledge Analysis), a product of the University of Waikato (New Zealand), along with the support offered in every step of the Data Mining process, with a good graphic interface, besides natively implementing many clustering algorithms, including SimpleKmeans \cite{Arthur07}, led to its choice as the platform where we conducted our experiments.\\

WEKA’s K-means version, either for Euclidean distance or for Manhattan distance, use the closest neighbor technique trying to diminish problems with nominal attributes. The general rule is that for two values of numerical attributes X and Y, the result of X-Y is used in the distance calculation. When the attribute is nominal, value 0 is attributed when X and Y are the same and 1 when they differ.

\subsection{Preprocessing and Clustering – Phase 1}
In a first phase, the dataset used in this paper come from a financial institution and represent the accounts movement over a period of three months. The most relevant tables in this dataset model are the transaction and register ones, with 14.5 million and 4.5 million lines, respectively. In the pre-process step, data was clustered by customer with numerical attributes that indicated the monthly average: of services used; transactions made; debit transactions made, credit transactions made. Besides, we included the average monetary value of these transactions. To each one of these attributes the standard deviation is also used, since there is a major variation between the minimum and maximum values. This table, called customers profiles resulted in 1.6 million lines.\\

Despite the already mentioned restrictions when using nominal attributes in cluster development with K-means, we decided to separate data and test both scenarios, numerical and nominal attributes. The motivation for this decision was the big variation between the minimum and maximum attribute values mentioned above, where the biggest variation stood between 0.01 minimum and 536,852,446.89 maximum.\\

\begin{figure}[!b]
\centering
\includegraphics[width=3.5in]{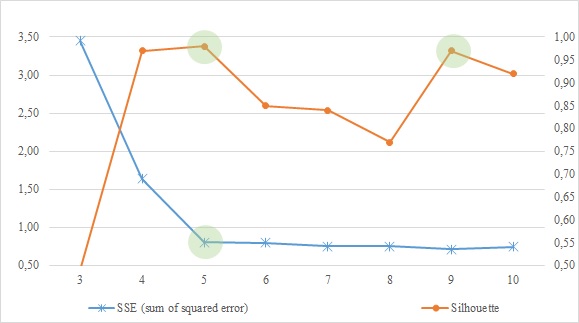}
\caption{Clusters evaluation (numeric attributes) - Phase 1}
\label{num_ph1}
\end{figure}

\begin{figure}[!t]
\centering
\includegraphics[width=3.5in]{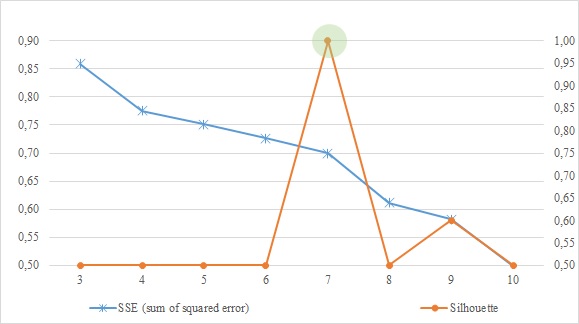}
\caption{Clusters evaluation (nominal attributes) - Phase 1}
\label{nom_ph1}
\end{figure}

As evaluation measures trying to define the initial number of clusters, we used the Silhouette Coefficient and the SSE (sum of squared error), which indicated the numbers of five and seven clusters. Fig. \ref{num_ph1} and Fig. \ref{nom_ph1} show the results.\\

In this case, the clusters created with nominal attributes showed a more coherent group of customers, for instance: old customers with a high use of services, high monetary values involved, credited values are rapidly withdraw (Cluster 1); customers’ account with less than 4 years old, high use of services, high quantity of entries, however, low monetary values and credited values rapidly withdrawn (Cluster 3).

\subsection{Rules Generation and Evaluation – Phase 1}
In the subsequent step of the process, rule generation, the PART algorithm was used, which is also included in the WEKA environment, and implements the C4.5 Decision Tree algorithm for interactions and uses the best leaf technique in rule generation \cite{Frank98}. In the executions all the default values suggested by WEKA were used.\\

\begin{figure}[!b]
\centering
\includegraphics[width=3.5in]{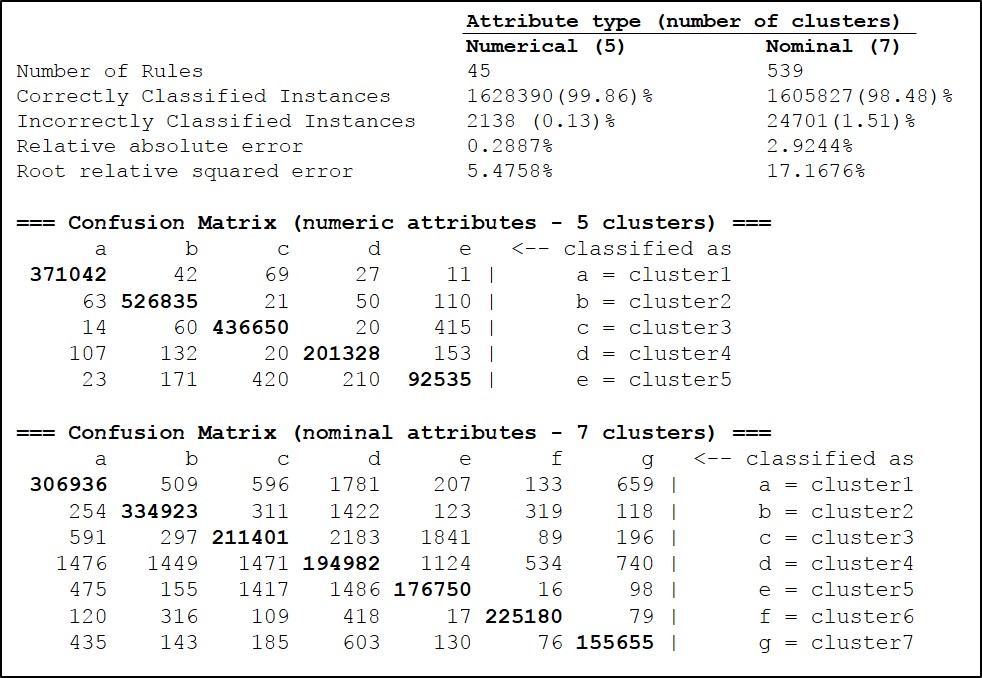}
\caption{Metrics of Rules Generation with Algorithm PART – Phase 1}
\label{part_ph1}
\end{figure}

Fig. \ref{part_ph1} shows that, in this case, the metrics have had an inverse result, meaning that the clusters based on numerical attributes had better results. Nevertheless, the rules generated, in both cases, demonstrate not being appropriate for the research goal.\\

\begin{figure}[!t]
\centering
\includegraphics[width=3.5in]{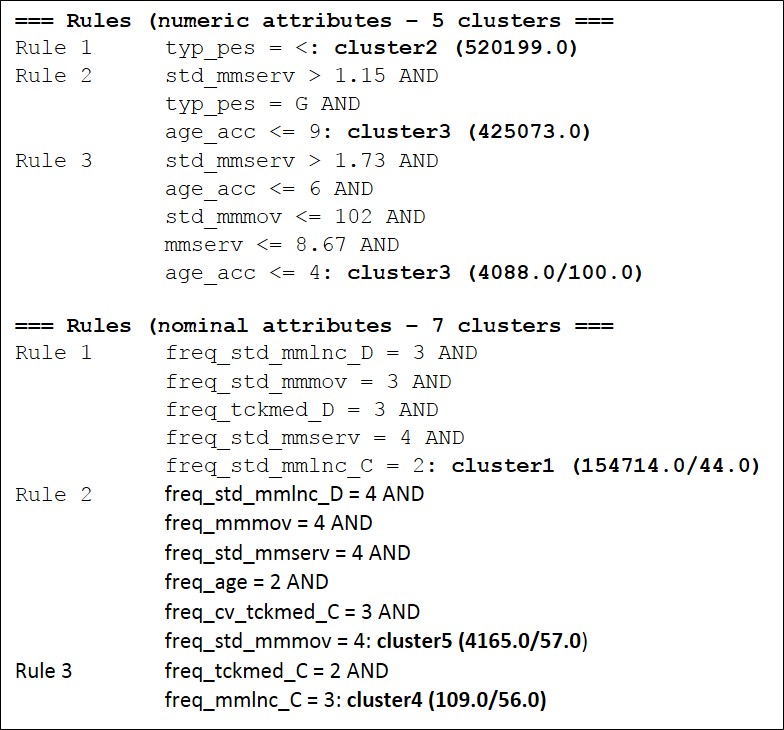}
\caption{Rules Generated by Algorithm PART – Phase 1}
\label{rules_ph1}
\end{figure}

Fig. \ref{rules_ph1} shows a few examples of rules with little use for the proposed study, whether it be for its simplicity and incapacity in helping taking a decision; or for its complexity, but resulting in few examples, also becoming disposable.

\subsection{Preprocessing and Clustering – Phase 2}
Considering that the obtained results were unsatisfactory, a new strategy was adopted. This consisted of creating a new profile table; use numerical attributes, but more directed towards the research goal; use a more ample database; make experiments with other rule generating algorithms.\\

Information with the monetary value involved in the transaction, financial transfers between banks and temporality of incoming and outgoing financial resources, started being part of the new customer profile table. The new checking accounts database incorporated transactions of the whole year of 2014 and the main tables, transactions and register, started to have 90.6 million and 5.1 million lines, respectively.\\

The compliance analysts from the financial institution that provided the database, besides confirming the importance of the defined attributes, identified the transactions that have no connection with money laundering. For example, charges made by the bank. With this definition the quantity of lines in the table will be reduced and the clusters generated will be more specialized on transactions that might actually correspond to money-laundering operations.\\

\begin{figure}[!t]
\centering
\includegraphics[width=3.5in]{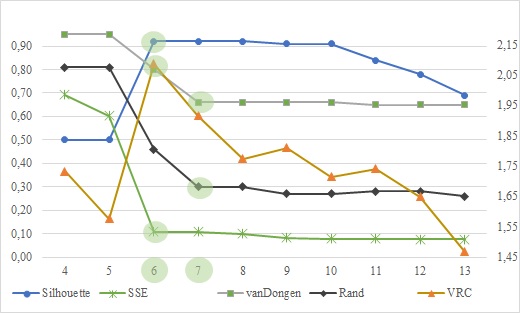}
\caption{Clusters evaluation (numeric attributes) - Phase 2}
\label{num_ph2}
\end{figure}

The new customer profile table remained with 2.4 million lines after the clustering and removal of the insignificant transactions. In the search of the adequate number of cluster to be used the SimpleKmeans was executed 10 times, the Silhouette Coefficient \cite{Rousseeuw87}, SSE, VRC (variance ratio criterion) \cite{Calinski74}, Van Dongen and Rand \cite{Silke07} metrics were analyzed, and the values found. Fig. \ref{num_ph2} shows these results.\\

The Silhouette metrics, SSE and VRC indicate six clusters as the ideal number, while Van Dongen and Rand indicate seven clusters. Both Silhouette and SSE show a stability line starting from number six, which can corroborate number seven, identified by the other metrics. However, the standard procedure is to make the choice on the curve “elbow” or the higher value, depending on the metric.\\ 
 
All the algorithm executions were made using the database Split function in the proportion of 66\% for training and 34\% for testing. The six clusters generated show excellent customer grouping, allowing identifications as:\\

Cluster \#3 – Standard Customer: biggest group of customers with high use of services, transactions financial values indicating intermediate customers. The money flows into account and the following days is withdraw;\\

Cluster \#4 – Group of Risk: high quantity of transactions, with low use of services; low financial values; money flows in and the same day or in a small amount of time is transferred to another financial institution.
The small difference between six and seven for the suggested quantity of clusters indicates the need to verify the result with 7 clusters. The redistribution of instances for the creation of the seventh cluster didn’t affect the basic characteristics of the first six clusters. In proportional terms Cluster \#3 was the one that most gave elements for the creation of Cluster \#7, which the characteristics may be defined as follows:\\

Cluster \#7 – Group of Risk 2: older accounts profile with great use of services and great volume of transactions. Financial values concentrated on areas called “legal limits”. A bigger percentage of outgoing financial resources, although with a low transference rate to other institutions. High rate of transfers between accounts of the same institution.\\

\begin{table}[!b] 
\caption{DISTRIBUTION OF INSTANCES BY CLUSTER} 
\begin{tabular}{c} 
\includegraphics[width=3.5in]{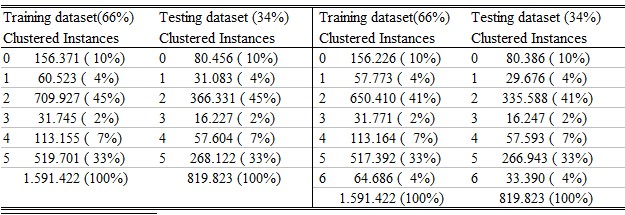}\\ 
\end{tabular} 
\label{inst}
\end{table} 

Because of the characteristics presented by Cluster \#7, its maintenance is important in the system configuration, thus, we started working with the creation of seven clusters. Tab. \ref{inst} shows that this choice doesn’t change the quality of the result.\\

\begin{figure}[!t]
\centering
\includegraphics[width=3.5in]{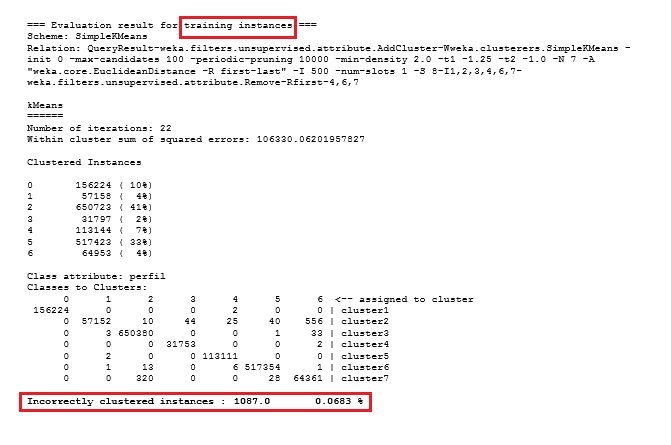}
\caption{Clusters Assessment Report (training instances)}
\label{training}
\end{figure}

\begin{figure}[!b]
\centering
\includegraphics[width=3.5in]{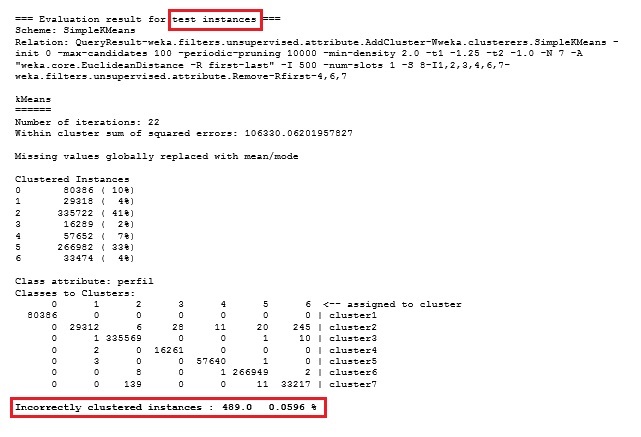}
\caption{Clusters Assessment Report (test instances)}
\label{test}
\end{figure}

Executing the algorithm with the generated cluster evaluation function we obtained a confusion matrix with a level of accuracy above 99\% if we consider the incorrect classification rates of 0.0683\% e 0.0596\%, for the training base and testing base respectively, as shown in Fig. \ref{training} and Fig. \ref{test}.\\

For rule generation, PART algorithm experiments were made, J48 \cite{Quinlan93} and JRip \cite{Cohen95}, using the WEKA tool default parameters and restraining the rule coverage to 100 and 1000 instances. To each of these options the PART algorithms and J48 were also executed with the “reducedErrorPruning” option activated, in the JRip algorithm this option is already part of the implementation.\\

\begin{figure}[!t]
\centering
\includegraphics[width=3.5in]{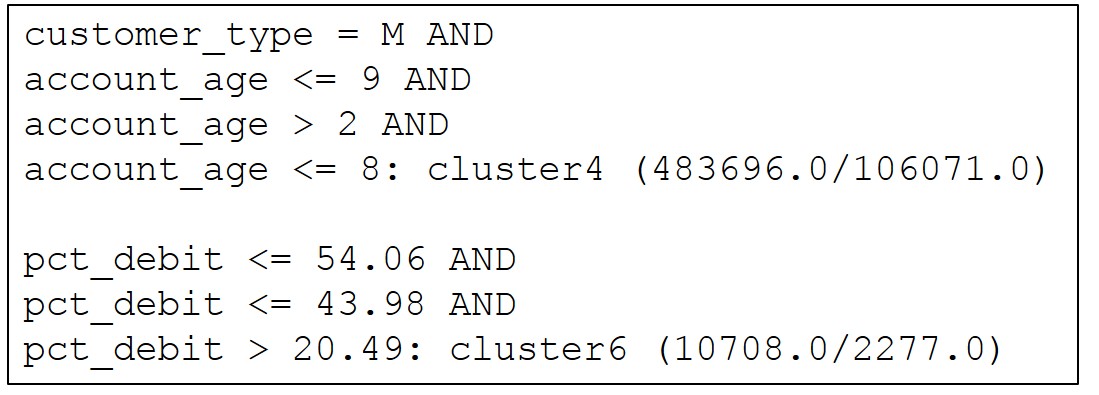}
\caption{Rules Generated by Algorithm PART – Phase 2}
\label{part_ph2}
\end{figure}

This configuration (15 experiments) was executed twice: one using database Split in the proportion of 66\% for training and 34\% for testing; and another with cross-validation (10 folds). Tab. 2, given in the Appendix, shows the result of the 30 executions. 
Fig. \ref{part_ph2} present the rules generated that also don’t have the expected quality, with strange repetitions of attributes or attributes conflicts.\\

Following these results we decided to separate the profiles, now with class attribute, and perform the tests again. The profiles were separated in three value groups with equivalent quantity of instances percentage, if possible. Some attributes, because of a concentration of occurrences of one particular value, were divided in only two groups. The experiments were performed executing the same algorithms and using the same parameters of the previous experiment. Tab. 3, given in the Appendix, presents this results.\\ 

The J48 and JPART algorithms present the best results in the group of experiments using Split and cross-validation, respectively. In both cases the number 1.000 limited the minimum instances per rule for PART and the minimum of leaves on the J48 tree. Although not being significant, the indicators presented better results if compared with previous experiments.\\

To improve the experiment with the purpose of verifying the behavior of these indicators, PART and J48 algorithms were executed 22 times each, varying the minimum number of instances by rule and of leaves in the tree. The minimal number is each algorithm’s default and the maximal number is the size of the smallest cluster generated (2 – 40,000).\\

\begin{figure}[!b]
\centering
\includegraphics[width=3.5in]{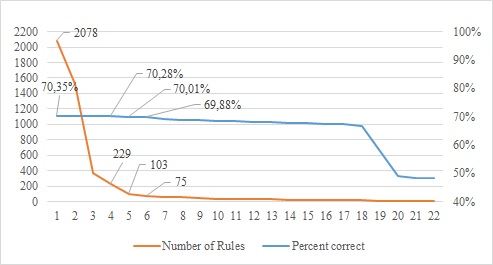}
\caption{Algorithm PART (Number of Rules \& Percent Correct)}
\label{part_nr}
\end{figure}

\begin{figure}[!t]
\centering
\includegraphics[width=3.5in]{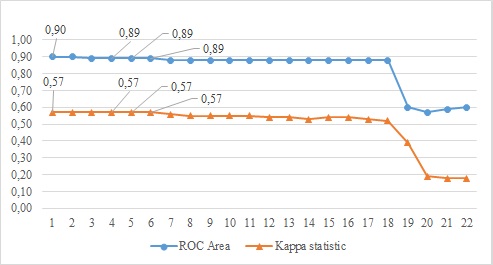}
\caption{Algorithm PART (ROC Area \& Kappa Statistic)}
\label{part_roc}
\end{figure}

\begin{figure}[!t]
\centering
\includegraphics[width=3.5in]{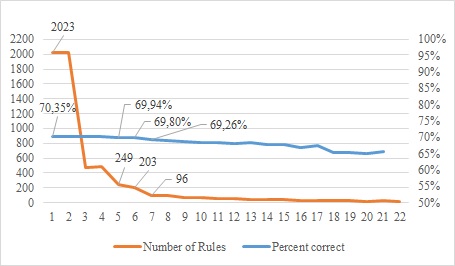}
\caption{Algorithm J48 (Number of Rules \& Percent Correct)}
\label{j48_nr}
\end{figure}

\begin{figure}[!t]
\centering
\includegraphics[width=3.5in]{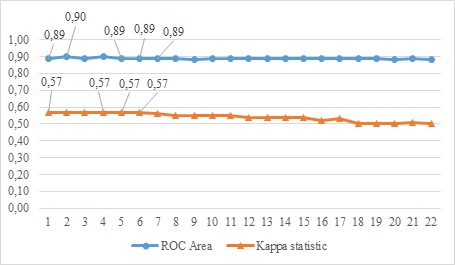}
\caption{Algorithm J48 (ROC Area \& Kappa Statistic)}
\label{j48_roc}
\end{figure}

Figures \ref{part_nr} to \ref{j48_roc} show the experiment results and demonstrate that the measure of rule quality is reversibly proportional to the increase of the minimum quantity of instances for the coverage of rules or for the leaves of the tree.\\

\begin{figure}[!t]
\centering
\includegraphics[width=3.5in]{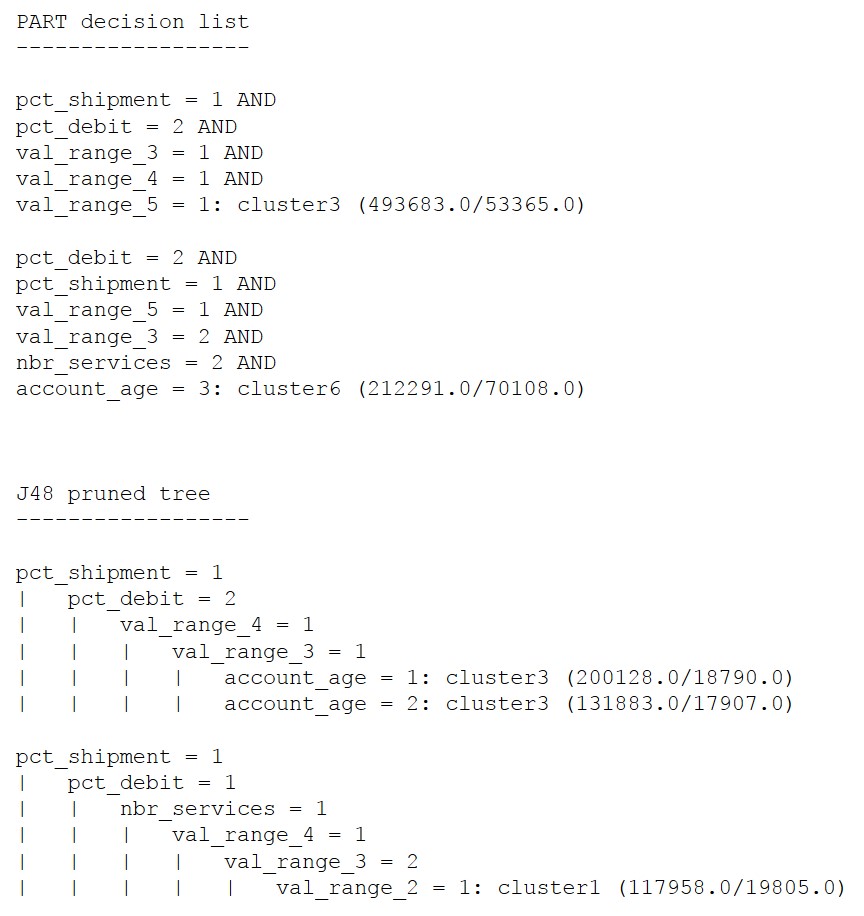}
\caption{Rules Generated by Algorithms PART and J48 – Phase 2}
\label{rules_ph2}
\end{figure}

With this experiment we understood that the results shown in Table 3 can be used for generating and analyzing rules. An analysis on rules generated by algorithms J48 and PART, using the parameters of the best result, makes it possible to notice an improvement in quality: more complete, without redundancies. Fig. \ref{rules_ph2} shows an example of the generated rules.

\section{Conclusion} 
In this work we presented the work on a learning component of an anti money-laundering system. The ultimate goal of our work is to have a tool that can assist financial institutions in the prevention and fight of money-laundering activities. We detailed the activities related to account movement databases in order to build client profiles, clusters, and the subsequent generations of rules that will be part of the intelligent agent’s knowledge bases, responsible for the identification of suspicious transactions.
 
The quest for general client profiles, allied to the use of a database that covered a small time span (3 months) produced good results for both the cluster evaluation metrics and generated rules. Nevertheless, precision was not as good as we expected.\\

The definition of client profiles that are more tailored to the system’s goal, with a database of a greater time span (1 year) and a more thorough exploration of the types of attributes available for the used algorithms, produced better results. It was possible to build clusters that represent risk groups and rules both more discriminating and with a wider coverage, regarding the profiles’ attributes. This improvement was validated by a human specialist from the financial institution that provided the data.

\bibliographystyle{IEEEtran}
\bibliography{ClientProfiling2015}

\onecolumn \maketitle \vfill
\appendix 
\section{Appendix}

\begin{table}[!h] 
\caption{EXPERIMENT FOR GENERATION OF RULES (NUMERICAL ATTRIBUTES)}
\begin{center}
\begin{tabular}{c} 
\includegraphics[width=6.0in]{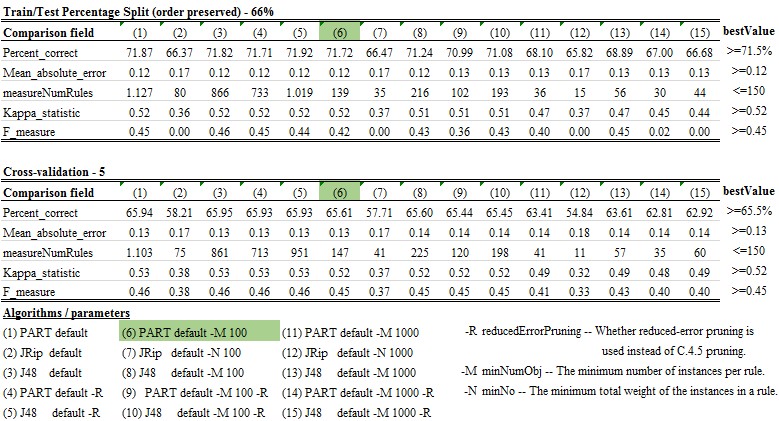} 
\end{tabular} 
\end{center}
\label{rules_num}
\end{table} 

\begin{table}[!h] 
\caption{EXPERIMENT FOR GENERATION OF RULES (NOMINAL ATTRIBUTES)} 
\begin{center}
\begin{tabular}{c} 
\includegraphics[width=6.0in]{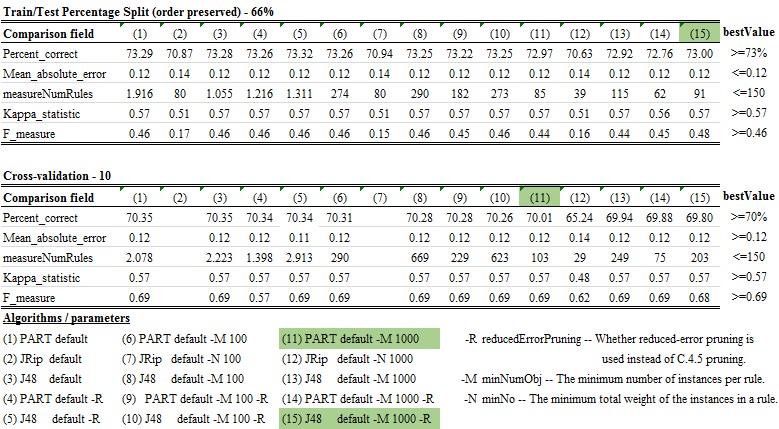}
\end{tabular} 
\end{center}
\label{rules_nom}
\end{table} 

\end{document}